\documentclass{ieeeaccess}
\usepackage{cite}
\usepackage{amsmath,amssymb,amsfonts}
\usepackage{algorithmic}
\usepackage{graphicx}
\usepackage{textcomp}
\usepackage{algorithm}

\usepackage{diagbox}
\def\BibTeX{{\rm B\kern-.05em{\sc i\kern-.025em b}\kern-.08em
    T\kern-.1667em\lower.7ex\hbox{E}\kern-.125emX}}
\begin{document}
\history{Date of publication xxxx 00, 0000, date of current version xxxx 00, 0000.}
\doi{10.1109/ACCESS.2017.DOI}

\title{A Hybrid Approach for Binary Classification of Imbalanced Data}
\author{\uppercase{Hsin-Han Tsai}\authorrefmark{1},
\uppercase{Ta-Wei Yang\authorrefmark{2}, Wai-Man Wong\authorrefmark{2}, and Cheng-Fu Chou}.\authorrefmark{1, 2, 3}
}
\address[1]{Department of Computer Science and Information Engineering, National Taiwan University, Taipei 10617, Taiwan, ROC.}
\address[2]{Graduate Institute of Networking and Multimedia, National Taiwan University, Taipei 10617, Taiwan, ROC.}
\address[3]{Information Technology Office, National Taiwan University Hospital, Taipei 100225, Taiwan, ROC.}
\tfootnote{This research was supported in part by the Ministry of Science and Technology of Taiwan (MOST 110-2622-8-002-018), National Taiwan University (NTU 109-3111-8-002-002), and Cathay Life Insurance.}

\markboth
{Author \headeretal: Preparation of Papers for IEEE TRANSACTIONS and JOURNALS}
{Author \headeretal: Preparation of Papers for IEEE TRANSACTIONS and JOURNALS}

\corresp{Corresponding author: Hsin-Han Tsai (e-mail: hhtsai@cmlab.csie.ntu.edu.tw).}

\begin{abstract}
Binary classification with an imbalanced dataset is challenging. Models
tend to consider all samples as belonging to the majority class. Although existing
solutions such as sampling methods, cost-sensitive methods, and ensemble
learning methods improve the poor accuracy of the minority class, these methods
are limited by overfitting problems or cost parameters that are 
difficult to decide. We propose HADR, a hybrid approach with dimension reduction 
that consists of data block construction, dimentionality reduction, and
ensemble learning with deep neural network classifiers. We evaluate the
performance on eight imbalanced public datasets in terms of recall, G-mean, and
AUC. The results show that our model outperforms state-of-the-art methods.
\end{abstract}


\titlepgskip=-15pt

\maketitle

\section{Introduction}
\label{sec:introduction}
\PARstart{B}{inary} classification of imbalanced data is a highly active topic
because it is a common scenario in the real world, for instance, credit card fraud
detection, defect detection, and rare disease diagnosis. The main feature of
imbalanced data is that most samples belong to a single class called the
\emph{majority class} and the rest of the samples belong to the other class, called the \emph{minority
class}. Typically, the minority class is more valuable than the majority class.
However, it is challenging to train a classifier which accurately distinguishes the minority
class. For example, consider a rare but fatal disease. Since it
is fatal, it is more important to discover a confirmed case than to find 
a healthy case. However, as the probability of a person having the disease is
very low, the trained classifier might tend to judge all cases as healthy
cases if it is trained for high accuracy.  Thus patients with the disease would
  tragically miss the opportunity to receive life-saving treatment.   
Hence, the challenge in training a classifier of imbalanced data is that the
classifier easily ignores misclassification of the minority class.

The literature has shown that traditional machine learning methods often fail
when applied to imbalanced data classification~\cite{ref:hybrid-rusboost-seiffert10}. 
Studies on improving the
performance of imbalanced data classification can be divided into three
categories: sampling methods, cost-sensitive methods, and ensemble learning methods. The
sampling method balances the number of instances between classes by generating
new samples or deleting existing samples, using techniques called 
\emph{oversampling}~\cite{ref:sampling-smote-chawla02} and \emph{undersampling}~\cite{ref:sampling-liu09},
respectively. Cost-sensitive methods improve the classifier by applying
different cost functions for misclassified samples~\cite{ref:cost-castro13}.
Ensemble learning is a general meta-method of machine learning, which seeks
better prediction performance by combining predictions from multiple
classifiers~\cite{ref:ensemble-alam18}. However, each method has
limitations which must be accounted for.
For sampling methods, undersampling may remove helpful information, and
oversampling may yield overfitted classifiers~\cite{ref:sampling-mease07}.
For cost-sensitive methods, it is not trivial to decide the cost parameters for
minority and majority classes. Also, the ensemble learning method may cause biased
results due to the unstable samples, which means the number of positive 
predictions is close to the number of negative predictions.  

Recently, a hybrid method called DDAE~\cite{ref:hybrid-ddae-yin20} has been 
proposed which integrates the advantages of the sampling,
cost-sensitive, and ensembling methods. However, if the imbalance ratio is
large and the amount of positive samples is small, the size of each data block
constructed by the method's data block construction (DBC) component is also small. The
data space improvement (DSI) component then easily overfits on each data block.
The following ensemble learning (EL) component based on the adaptive weighted
adjustment (AWA) component may fail if most of the unstable samples are positive.

In this paper, we propose a hybrid approach with dimension reduction (HADR)
for binary imbalanced classification problems 
that addresses the problems of the above categories. HADR contains three
components: (i)~a data block construction (DBC) component responsible
for dividing the input data into nearly balanced data blocks. 
(ii)~A dimensionality reduction (DR) component which reduces the dimensionality of
the features and improves the data space by using metric learning for kernel
regression. DR not only encodes the features but also prunes redundant
features. In this step, we seek to mitigate the overfitting problem of
DDAE. (iii)~An ensemble learning component which combines multiple classifiers
via a voting mechanism. We apply multilayer perceptrons (MLPs) as
classifiers because of their low bias. High variance can be reduced
by the bagging ensemble method. We evaluate the performance on eight
imbalanced public datasets in terms of the recall, G-mean, and AUC metrics. The results
show that
HADR outperforms state-of-the-art methods. In particular, in the Wine3vs5
dataset, the recall of HADR is 
66.7\% better than that of DDAE.   

We summarize the rest of this paper as follows: we discuss related works in 
Section~\ref{sec:related}, the proposed methodology in Section~\ref{sec:method}, and
the experiments in Section~\ref{sec:exp}. We conclude in Section~\ref{sec:conclusion}.

\section{Related Work}
\label{sec:related}
Our DR component uses metric learning to reduce the dimensionality of the
dataset to find an effective data representation. Studies on data
representation apply techniques such as metric 
learning~\cite{ref:metric-lin19,ref:metric-wang18} or data space 
improvement~\cite{ref:hybrid-ddae-yin20}. 
Several techniques can be used to address data imbalance. Based on the
different balancing techniques for data imbalance, we classify the most
relevant studies into three types: sampling
methods~\cite{ref:sampling-smote-chawla02,ref:sampling-mease07,ref:sampling-he09,ref:sampling-liu09,ref:sampling-mwmote-barua14,ref:sampling-racog-das14},
cost-sensitive
methods~\cite{ref:cost-elkan01,ref:cost-sun07,ref:cost-thai10,ref:cost-castro13,ref:cost-nikolaou16,ref:cost-dumpala18},
and ensemble learning
methods~\cite{ref:ensemble-breiman96,ref:ensemble-hastie01,ref:ensemble-alam18,ref:ensemble-fang19}. 
Each method effectively alleviates the influence of imbalanced data.
Multiple methods may also be combined for further
improvements~\cite{ref:hybrid-rusboost-seiffert10,ref:hybrid-jing21,ref:hybrid-ddae-yin20}.
In recent years, due to the growth of deep learning, some components (sampling, cost-sensitive, ensemble, and data
representation components) have been replaced with neural 
networks~\cite{ref:deep-huang16,ref:deep-dablain21,ref:deep-zhang20}.

In the following subsections, we will review sampling,
cost-sensitive, and ensemble methods. We will also discuss the
state-of-the-art DDAE model~\cite{ref:hybrid-ddae-yin20}.

    \subsection{Sampling methods}
	 The purpose of a sampling method is to turn imbalanced data into 
	 balanced data either by generating new samples or deleting existing
	 samples. Hence sampling methods fall into two
	 categories: undersampling and oversampling.

    \subsubsection{Undersampling}
	 The concept of undersampling is to remove majority samples to
	 balance the data. Selection approaches include
	 random undersampling and informed undersampling. 
	 Random undersampling techniques~\cite{ref:sampling-he09} involve randomly
	 selecting examples from the majority class and removing them from the training
	 data. The limitation with this technique is the removal of samples
	 without accounting for their usefulness or importance while determining the
	 decision boundary between the classes. That is, we lose 
	 valuable information with this technique. To account for this,
	 Liu et~al.~\cite{ref:sampling-liu09} propose informed undersampling. They sample
	 the majority data into several subsets and train a classifier for each
	 subset. Then they combine the classifiers to produce the final
	 prediction. This makes better use of the majority class than random
	 undersampling.

    \subsubsection{Oversampling}
	 Oversampling duplicates or generates sufficient
	 minority samples to achieve a balanced state. Here we also have two main
	 approaches: random oversampling and synthetic oversampling. In random
	 oversampling, we randomly select and duplicate minority samples until
	 the data is balanced. Then we train the classifier on the derived
	 balanced data. Although this prevents the classifier from predicting all samples
	 as the majority class, such random oversampling often results in data 
	 overfitting, as the prediction of the minority
	 class relies on the information in the minority samples. 
	 The synthetic minority oversampling technique 
	 (SMOTE)~\cite{ref:sampling-smote-chawla02} addresses this limitation by generating synthetic samples
	 from minority classes. The idea of SMOTE is to apply the k-nearest
	 neighbors algorithm, choose $n$ samples, and then generate the
	 synthetic sample between the sample in question and its neighbors. SMOTE
	 prompted a number of related studies~\cite{ref:sampling-mwmote-barua14,ref:deep-dablain21}.
	 Alam et~al.~\cite{ref:ensemble-alam18} propose a new method for addressing
	 multi-class imbalance which differs from other data
	 balancing methods like sampling and underbagging. The basic idea is to
	 partition data from multi-class imbalanced problems into several balanced
	 problems. This is a unique recursion-based approach that partitions the
	 imbalanced data into balanced data.
    
    \subsection{Cost-sensitive methods}
	 Cost-sensitive methods help to improve the classifier by applying
	 different costs to different misclassified 
	 samples~\cite{ref:cost-castro13}. First, we review the cost matrix in
	 Table~\ref{tab:cost-matrix}. Following the notation in
	 \cite{ref:cost-elkan01}, let $C(i,j)$ be the cost when class~$i$ is predicted
	 and class~$j$ is the ground truth. Thus $C(0,0)$, $C(0,1)$,
	 $C(1,0)$, and $C(1,1)$ are true negative (TN), false negative (FN),
	 false positive (FP), and true positive (TP), respectively. The optimal
	 classifier can be trained by minimizing the cost function
    \begin{equation}\label{fun:cost}
        L(x,i)=\sum_j P(j|x)C(i,j).
    \end{equation}
	 It is intuitive to set the costs in FN and FP to be greater than TN and TP.
	 When optimizing (\ref{fun:cost}), the classifier learns to focus on 
	 false predicted samples. However, it is still a challenge to define the
	 cost matrix. One heuristic is to assign costs based on the inverse class
	 distribution. For example, if we have an imbalance ratio of $100:1$ between the
	 majority and minority classes, we can set $C(0,0)=0$, $C(0,1)=100$,
	 $C(1,0)=1$, and $C(1,1)=0$. This will force the classifier to focus on correctly
	 predicting the minority class. However, this heuristic assumes the class
	 distribution of the training data is equal to that of the test data. Another simple
	 way is to use a machine-learning model that predicts the probability of
	 each class, and combine this with a line search on a threshold, where samples are
	 assigned to every clear class label, to ultimately minimize the cost of
	 misclassification. The cost-sensitive method is a commonly-used method,
	 despite the difficulty in obtaining precise
	 misclassification cost parameters for minority and majority classes.

    \begin{table}[!t]
    \caption{Cost matrix\label{tab:cost-matrix}}
    \centering
    \begin{tabular}{|c|c|c|}
    \hline
     & Actual negative & Active positive\\
    \hline
    Predicted negative & $C(0,0)$ & $C(0,1)$\\
    \hline
    Predicted positive & $C(1,0)$ & $C(1,1)$\\
    \hline
    \end{tabular}
    \end{table}
    
    \subsection{Ensemble methods}
	 Ensemble learning combines the results from several classifiers to
	 improve prediction performance~\cite{ref:ensemble-hastie01}. The bagging
	 (bootstrap aggregating) algorithm~\cite{ref:ensemble-breiman96} and the
	 boosting algorithm are two main ensemble learning methods. One 
	 difference between the bagging algorithm and the boosting algorithm is
	 that boosting aggregates the results using weights. Given the effectiveness of ensemble
	 learning, many studies combine data
	 partitioning with ensemble learning for imbalanced 
	 classification~\cite{ref:ensemble-alam18}. 

    \subsection{DDAE}
	 DDAE~\cite{ref:hybrid-ddae-yin20} is a novel model based on k-nearest-neighbor 
	 classifiers for imbalanced data classification. 
	 DDAE contains four main components: (i)~a data block construction (DBC)
	 component responsible for dividing the input data into nearly
	 balanced data blocks; (ii)~a data space improvement (DSI) component which
	 brings samples in the same class closer together and further separates
	 samples from different classes; (iii)~an adaptive
	 weight adjustment (AWA) component which adjusts the ensemble learning weight 
	 for each classifier; and (iv)~an ensemble learning (EL)
	 component which combines multiple base classifiers via the use of weighted
	 voting.

\section{Method}
\label{sec:method}
    \subsection{Model Description}
	 Figure~\ref{fig:model} depicts the HADR architecture. There are three components
	 in the training phase: (i)~a data block construction (DBC) component,
	 which divides the training set into nearly balanced data blocks; (ii)~a
	 dimensionality reduction (DR) component, which reduces the dimensionality
	 of the data and improves the data space by using metric learning for kernel
	 regression; and (iii)~an ensemble learning (EL) component, which combines multiple
	 classifiers via a voting mechanism. In the testing phase, we directly
	 apply the trained DR component to the test set and then combine the
	 predictions of the trained classifiers.

    \Figure[t!][width=0.45\textwidth]{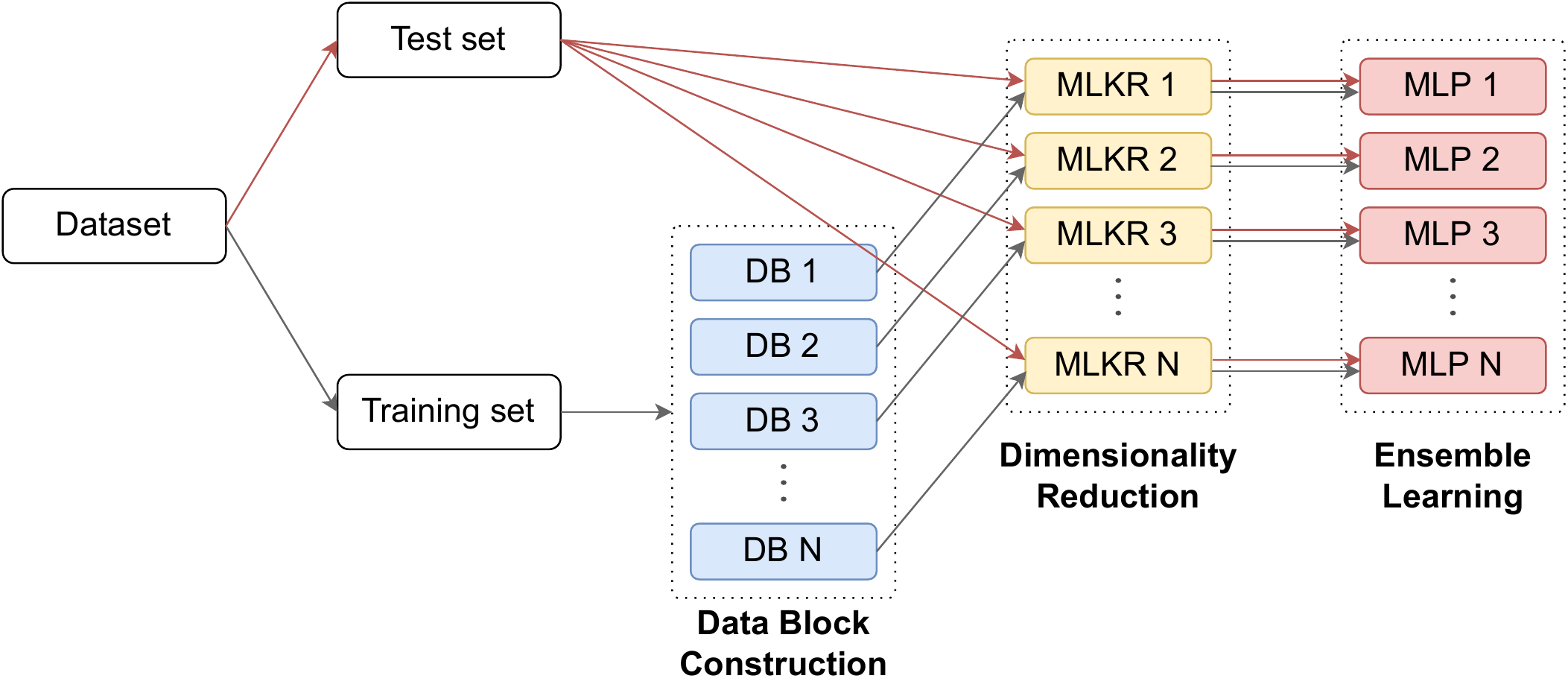}
	 {HADR architecture. During training, the training set is divided
	 into $N$ nearly balanced data blocks, and $N$ MLKR components are trained for
	 dimension reduction. After $N$ MLPs are trained for classification, 
	 the predictions of all the classifiers are combined via the EL component. During testing, we
	 directly apply dimension reduction to the test data and then combine the
	 predictions of the MLP classifiers.\label{fig:model}}

    \subsection{Data Block Construction (DBC) component}
	 To generate balanced data blocks, we divide the training data
	 into multiple blocks by partitioning the 
	 data~\cite{ref:ensemble-alam18}. We also modify the partitioning 
	 method~\cite{ref:ensemble-alam18} to make it more suitable for binary
	 classification. Let $S_{\mathit{min}}$ and $S_{\mathit{maj}}$ be the set of minority and
	 majority samples, respectively. Let $N_{\mathit{min}}$ and $N_{\mathit{maj}}$ be the number
	 of samples in $S_{\mathit{min}}$ and $S_{\mathit{maj}}$, respectively.  
	 Let $\mathit{ir}=N_{\mathit{maj}}/N_{\mathit{min}}$ be the imbalance ratio of the dataset. 
	 Algorithm~\ref{alg:dbc} shows the data block construction process. In the algorithm,
	 $\mathit{ir}^*$ can be either $ \lfloor \mathit{ir} \rfloor$ or $\lceil  \mathit{ir} \rceil. $
    
    \begin{algorithm}[h]
      \caption{Data Block Construction}
      \label{alg:dbc}
      \begin{algorithmic}[1]
        \REQUIRE
          Dataset $D$
        \ENSURE
          A set $B$ of $\mathit{ir}^*$ data blocks
        \STATE Obtain $S_{\mathit{min}}$, $S_{\mathit{maj}}$, $N_{\mathit{min}}$, and $N_{\mathit{maj}}$ based on $D$;
        \STATE Divide $S_{\mathit{maj}}$ into $\mathit{ir}^*$ chunks $\{C_1,C_2,\dots,C_{\mathit{ir}^*}\}$:
        \FOR{$i=1$ to $\mathit{ir}^*$}
          \STATE $B_i = C_i \cup S_{\mathit{min}}$;
        \ENDFOR\\
        \RETURN $B=\{B_1,B_2,\dots,B_{\mathit{ir}^*}\}$
      \end{algorithmic}
    \end{algorithm}

	 Figure~\ref{fig:dbc} illustrates Algorithm~\ref{alg:dbc}. First, we
	 obtain $S_{\mathit{maj}}$, $S_{\mathit{min}}$, $N_{\mathit{maj}}$, and $N_{\mathit{min}}$ according to their
	 classes. Then we obtain the imbalance ratio $\mathit{ir}=N_{\mathit{maj}}/N_{\mathit{min}}$ and split
	 $S_{\mathit{maj}}$ into $\mathit{ir}$ chunks $\{C_1, C_2, \dots, C_{\mathit{ir}}\}$. Since we
	 seek to maintain the balance in each data block $B_i$, $i=1,\dots,\mathit{ir}$,
	 we combine every $C_i$ with the minority data $S_{\mathit{min}}$ and form $\mathit{ir}$
	 nearly balanced data blocks $\{B_1, B_2, \dots, B_{\mathit{ir}}\}$.
    
    \Figure[t!][width=0.4\textwidth]{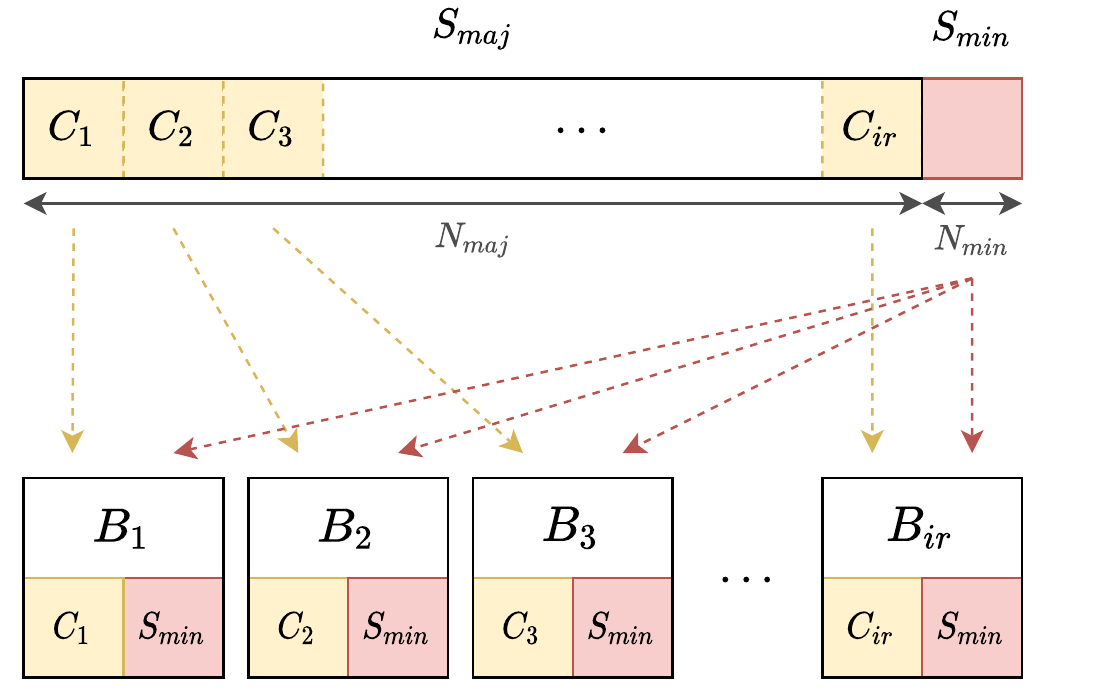}
	 {Data block construction. Suppose the imbalance ratio is $\mathit{ir}$. First,
	 divide the majority set ($S_{\mathit{maj}}$) into $\mathit{ir}$ chunks ($C_1, C_2, \dots,
	 C_{\mathit{ir}}$) and then combine each chunk with the minority set $S_{\mathit{min}}$ and form
	 a data block $B_i$. This yields \(\mathit{ir}\) data blocks.\label{fig:dbc}}

    \subsection{Dimensionality Reduction (DR) Component}
	 In machine learning or deep learning, dimensionality reduction is an
	 important step which extracts useful features and decreases the effect of
	 noise. Principal component analysis (PCA)~\cite{ref:pca-pearson01} is a
	 well-known dimensionality reduction technique. However, one 
	 limitation is that the projection is entirely unsupervised; thus side
	 information is sometimes ignored. Hence, we apply metric learning kernel 
	 regression (MLKR), another dimensionality reduction technique which 
	 can be viewed as a supervised PCA. 
    
	 We will demonstrate how PCA reduces the dimensionality of the data. Assume
	 that $\mathbf{x}_i\in \mathbb{R}^m$ is a sample with $m$~features.

    \subsubsection{PCA}
    First, we calculate the covariance matrix as
    \begin{equation}
    \label{eq:cov}
        \mathbf{C} = \sum_i \mathbf{x}_i \mathbf{x}_i^\top \in \mathbb{R}^{m\times m},
    \end{equation}
	 after which we compute eigenvalues $\lambda_j$ of \textbf{C} in descending
	 order and the corresponding eigenvectors $\mathbf{v}_j$ using
    \begin{equation}
    \label{eq:eig}
        \mathbf{C}\mathbf{v}_j = \lambda_j \mathbf{v}_j,
    \end{equation}
    where $\lambda_1 \geq \lambda_2 \geq \dots \geq \lambda_m$ and  $j=1,2, \dots, m$.
	 Then we project the sample $\mathbf{x}_i$ onto the lower dimensional space
	 and derive $\hat{\mathbf{x}}_i$ as 
    \begin{equation}
    \label{eq:reduc}
        \hat{\mathbf{x}}_i = \left[ \mathbf{v}_1^\top \mathbf{x}_i, \mathbf{v}_2^\top \mathbf{x}_i, \dots , \mathbf{v}_l^\top \mathbf{x}_i  \right]^\top,
    \end{equation}
    where $l \leq m$.
    
	 Due to the above-mentioned limitation with standard PCA, we apply MLKR instead. We
	 demonstrate it as follows.

    \subsubsection{MLKR}
	 In the proposed method, we apply MLKR~\cite{ref:mlkr-weinberger07} as the
	 tool for dimensionality reduction for each data block. MLKR 
	 learns a distance function $d(\cdot,\cdot)$ by minimizing the loss
	 function 
    \begin{equation}
    \label{eq:loss}
        \mathcal{L} = \sum_i (y_i - \hat{y_i})^2,
    \end{equation}
    where
    \begin{equation}
    \label{eq:yhat}
        \hat{y_i} = \frac{\sum_{j\neq i} y_jk_{ij}}{\sum_{j\neq i} k_{ij}}
    \end{equation}
    and
    \begin{equation}
    \label{eq:k}
        k_{ij} = \frac{1}{\sigma \sqrt{2\pi}} e^{-\frac{d(\mathbf{x}_i, \mathbf{x}_j)}{\sigma^2}}.
    \end{equation}
	 In (\ref{eq:loss}), $\hat{y_i}$ represents the approximated $y_i$ and is
	 calculated as the weighted average of the nearby $y_j$ in (\ref{eq:yhat}). The
	 kernel function $k_{ij}$ is usually defined as a Gaussian kernel in
	 (\ref{eq:k}). The distance function $d(\mathbf{x}_i, \mathbf{x}_j)$
	 measures the ``distance'' between two samples $\mathbf{x}_i$ and
	 $\mathbf{x}_j$. The distance function is defined as
    \begin{equation}
    \label{eq:d}
        d(\mathbf{x}_i, \mathbf{x}_j) = (\mathbf{x}_i - \mathbf{x}_j)^\top \mathbf{M} (\mathbf{x}_i - \mathbf{x}_j),
    \end{equation}
	 where the matrix $\mathbf{M}$ represents a Mahalanobis matrix, which is symmetric
	 positive semi-definite. That is, for any vector $\bf{v}$, $\mathbf{M}$ satisfies
	 $\mathbf{v}^\top \mathbf{M} \mathbf{v} \geq 0$. The distance function $d$ can 
	 be interpreted as a generalized Euclidean metric. It is indeed an
	 Euclidean metric if $\mathbf{M}$ is an identity matrix. The authors of
	 \cite{ref:mlkr-weinberger07} expect the matrix $\mathbf{M}$ to help $d$ to
	 measure the distance more appropriately. However, it is difficult to maintain
	 the positive semi-definite property in the training procedure. They solve
	 this problem by defining $\mathbf{M}=\mathbf{A}^\top \mathbf{A}$ which
	 transforms the learning target from the constrained $\mathbf{M}$ to the
	 unconstrained $\mathbf{A}$. Thus we rewrite (\ref{eq:d}) as
    \begin{equation}
    \label{eq:da}
        d(\bf{x_i}, \bf{x_j}) = \lVert \mathbf{A}(\bf{x_i} - \bf{x_j})\rVert^2.
    \end{equation}
	 After optimization of (\ref{eq:loss}), the learned $\mathbf{A}$ projects
	 the data $\mathbf{x}_i$ onto the embedding space (\ref{eq:mapping}),
	 yielding improved regression performance:
    \begin{equation}
    \label{eq:mapping}
        \mathbf{x}_i \to \mathbf{Ax}_i.
    \end{equation}

    \subsubsection{Supervised PCA}
	 In \cite{ref:mlkr-weinberger07}, the authors showed that MLKR also can be
	 interpreted as a supervised PCA. Assume that the covariance matrix $\mathbf{C}$
	 of the input $\mathbf{x}_i$ is an identity matrix (this can be accomplished
	 by whitening $\mathbf{x}_i$). Thus the covariance matrix of $\mathbf{Ax}_i$ is
	 $\mathbf{C^A}=\mathbf{AA^\top}$. Suppose $\gamma_j$ and $\mathbf{w}_j$ are
	 the eigenvalue and the corresponding eigenvector of $\mathbf{C^A}$, where
	 $\gamma_1 \geq \gamma_2 \geq \cdots \geq \gamma_m$.    
	 According to the PCA algorithm, we can
	 derive the reduced feature space by multiplying $\mathbf{x}_i$ by the
	 selected number of eigenvectors as 
    \begin{equation}
    \label{eq:w}
        \hat{\mathbf{x}}_i = \left[ \mathbf{w}_1^\top \mathbf{Ax}_i, \mathbf{w}_2^\top \mathbf{Ax}_i, \dots , \mathbf{w}_l^\top \mathbf{Ax}_i  \right]^\top \in \mathbb{R}^l,
    \end{equation} 
    which yields
    \begin{equation}
        \mathbf{C^Aw}_j = \mathbf{AA}^\top \mathbf{w}_j = \gamma_j \mathbf{w}_j.
    \end{equation}
    Multiplying by $\mathbf{A}^\top$ then yields
    \begin{align}
        \mathbf{A}^\top\mathbf{AA}^\top \mathbf{w}_j &= \gamma_j \mathbf{A}^\top\mathbf{w}_j,\\
        \mathbf{A}^\top\mathbf{A}(\mathbf{A}^\top \mathbf{w}_j) &= \gamma_j (\mathbf{A}^\top\mathbf{w}_j),\\
        \mathbf{M}(\mathbf{A}^\top \mathbf{w}_j) &= \gamma_j (\mathbf{A}^\top\mathbf{w}_j).
        \label{eq:u}
    \end{align}
	 Let $\mathbf{u}_i$ be the eigenvector of $\mathbf{M}$. From (\ref{eq:u}),
	 we have $\mathbf{u}_i = \mathbf{A}^\top\mathbf{w}_j$. Thus, we rewrite
	 (\ref{eq:w}) to derive
    \begin{equation}
    \label{eq:rewritew}
        \hat{\mathbf{x}}_i = \left[ \mathbf{u}_1^\top \mathbf{x}_i, \mathbf{u}_2^\top \mathbf{x}_i, \dots , \mathbf{u}_l^\top \mathbf{x}_i  \right]^\top \in \mathbb{R}^l.
    \end{equation} 
	 Equations~(\ref{eq:u}) and~(\ref{eq:rewritew}) show that since $\mathbf{M}$ is
	 actually the covariance matrix of $\mathbf{x}_i$, we can directly
	 apply the PCA algorithm. Moreover, $\mathbf{M}$ is learned in such a way that it can be
	 interpreted as a supervised form of PCA given input data $\mathbf{x}_i$. That is, it
	 does not waste label information in the input data.
    
	 In the proposed method, DR is an important step since the DBC component may
	 create multiple small data blocks if the positive samples are rare, 
	 in which case the classifiers are likely to overfit the data blocks.  
	 Moreover, the DSI component in DDAE may fail due to such
	 outliers. DR by MLKR addresses this problem not only by reducing the
	 dimensionality by pruning redundant features, but also by selecting 
	 important features via feature space encoding.
    
    \subsection{Ensemble learning}
	 Ensemble learning, the last component of the proposed model architecture,
	 combines the classifiers to achieve better results. In the
	 previous component, we obtain blocks of features whose dimensionality is
	 reduced. Then we utilize those features as inputs for classifiers. Each
	 classifier is an independent multilayer perceptron (MLP) with the following
	 design. The MLP contains two hidden layers (including the input layer) and
	 a sigmoid function. Each hidden layer contains 10 nodes. Figure~\ref{fig:mlp}
	 shows the structure of the MLP classifier.

    \Figure[t!][width=0.4\textwidth]{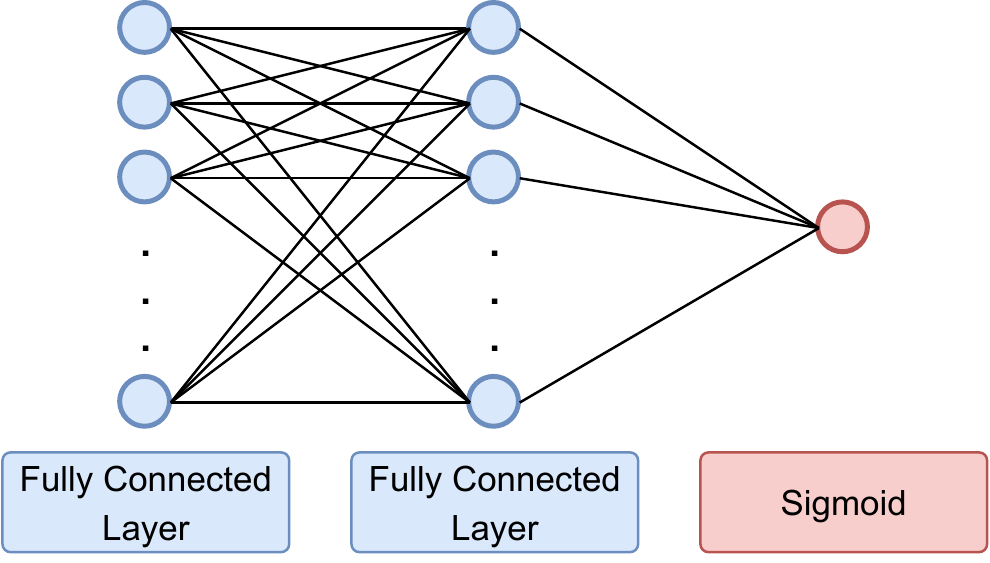}
    {Base classifier: multilayer perceptron \label{fig:mlp}}

	 The additional settings are described as follows. We use 1000 epochs
	 with a batch size of 10 and binary cross-entropy as the loss function.
	 After training all the MLP classifiers, we apply the majority voting mechanism.
    
	 In deep learning, MLPs are usually applied as classifiers. They are easy to
	 use while being sufficiently complicated to approach the actual
	 classifier. MLP exhibits low bias but high variance. We use bagging
	 ensemble methods like majority voting to average the variance and derive 
	 low-bias, low-variance results.
    
\section{Experiments}
\label{sec:exp}
In this section we will introduce the experimental setup and evaluate the
proposed model on eight imbalanced public datasets. We will show that HADR 
outperforms  state-of-the-art models.

    \subsection{Evaluation Metrics}
	 For an imbalanced classification model, we typically do not measure 
	 performance using accuracy. As an example, consider a dataset that contains one cancerous
	 patient and 99 non-cancerous patients. The model achieves $99\%$ accuracy
	 if all patients receive negative predictions. However, we have a 
	 greater interest in correctly predicting positive cases. In this case, the sensitivity
	 (recall) equals zero, so the model is essentially useless even if it has extremely
	 high accuracy. In our experiments, we apply recall, 
	 G-mean~\cite{ref:gmean-kubat97}, and ROCAUC which consider samples from both
	 majority and minority classes. To calculate these three metrics, we
	 must first derive the true positives (TP), false negatives (FN), false positives
	 (FP), and true negatives (TN) from the ground truths and predictions.

    \begin{enumerate}
	 \item \(\mathit{Recall}=\frac{\mathit{TP}}{\mathit{TP}+\mathit{FN}}\), 
	 which also called the true positive rate
	 (TPR) or sensitivity in the medical domain. The denominator represents the
	 number of all positive samples and the numerator represents the number of
	 positive samples that are correctly predicted. 
	 Recall focuses on the minority class.
	 \item \(\mathit{G}$-$\mathit{mean}= \sqrt{\mathit{Recall}*\mathit{TNR}}\), where \(\mathit{TNR}=\frac{\mathit{TN}}{\mathit{TN}+\mathit{FP}}\) denotes
	 the true negative rate. We have mentioned that recall is also called TPR. 
	 Thus G-mean is actually the product of TPR and TNR, which implies that this metric focuses
	 on measuring the balance between majority class accuracy and minority class
	 accuracy. In the previous example, if the model predicts all samples as
	 negative, then the model has a high TNR but a low TPR, which translates to a 
	 low G-mean.
    
	 \item AUC, which refers to the area under the ROC curve. This is a number
	 between zero and one. A higher AUC implies better performance. For the ROC
	 curve, the x-axis represents the false positive rate (FPR) while the y-axis
	 represents the TPR. To calculate AUC, we retrieve the predicted score from
	 the model. This is usually a probability, but it need not be. Given this
	 score, the final class prediction is based on a given threshold. Using
	 different thresholds, we derive different TPR-FPR pairs which we then use to
	 plot the ROC curve. The best
	 performance occurs with a 0\% FPR and a 100\% TPR, which means AUC equals one.  
	 Thus a larger AUC means we have a higher probability to get a high TPR and a low FPR
	 simultaneously. As described above, AUC simultaneously considers the performance of the majority
	 class and the minority class. Hence, it is an appropriate metric to
	 measure the classification of imbalanced data.
    \end{enumerate}
    
    \subsection{Experimental setup}
	 All experimental results were based on the eight imbalanced datasets. These
	 include Pc1, Pc3, Pc4, and Mw1 from OpenML~\cite{ref:openml-vanschoren14}, 
	 all open datasets for the detection of software defects. 
	 The other datasets are from the KEEL repository~\cite{ref:keel-alcal09}, 
	 where Wine3vs5 is used for wine quality
	 prediction, and the abalone datasets are used to predict the age of
	 abalone. Table~\ref{tab:tab1} shows the total number of samples, the number of
	 features in each sample, and the imbalance ratio (majority/minority) of the 
	 eight datasets. We randomly split each dataset into two parts for all
	 experiments: a training set (70\%) and a testing set (30\%). In the ensemble
	 learning component for each MLP block, we further randomly split the
	 training set into two parts: 80\% for MLP training and 20\% for
	 validation. Due to the lack of open-source code, we compared the proposed method
	 (HADR) with other methods following \cite{ref:hybrid-ddae-yin20}. In
	 addition to the comparison with the DDAE model, we also tested on other
	 state-of-the-art models: 

    \begin{table}[htbp]
	 \caption{Dataset summary. As some datasets have multiple classes,
	 ``vs'' means the binary classification of two selected classes.
	 ``Wine3vs5'' is short for the ``winequalityred3vs5'' dataset and
	 ``abalone20'' is short for the ``abalone20vs8910'' dataset.}
    \begin{center}
    \begin{tabular}{|c|c|c|c|}
    \hline
    Dataset & Samples & Features & Imbalance ratio \\
    \hline
    Pc1 &	1109 & 21 &  13.4\\
    Pc3	& 1563 & 37 &   8.8\\
    Pc4	& 1458 & 37 &   7.2\\
    Mw1	& 403 & 37 &    12.0\\
    Wine3vs5 & 691 & 11 & 68.1\\
    abalone9vs18 & 731 & 8 & 16.4\\
    abalone19 & 4174 & 8 & 129.4\\
    abalone20 & 1916 & 8 &   72.7\\
    \hline
    \end{tabular}
    \label{tab:tab1}
    \end{center}
    \end{table}
    
    \begin{itemize}
	 \item IML \cite{ref:metric-wang18}, which combines iterative metric
	 learning for the construction of a stable data space and the k-nearest
	 neighbors algorithm as a classifier.  
	 \item RP \cite{ref:ensemble-alam18}, which combines data partitioning for
	 blocks generation and ensemble learning, with majority voting rules. 
	 \item CAdaMEC \cite{ref:cost-nikolaou16}, a methodology that calibrates 
	 cost-sensitive learning methods and AdaMEC~\cite{ref:adamec-ting00}, which uses a
	 decision tree as a classifier.  
	 \item MWMOTE \cite{ref:sampling-mwmote-barua14}, a synthetic
	 oversampling method that weights minority class samples based on
	 the Euclidean distance to the nearest majority sample. It uses the
	 k-nearest neighbors algorithm as a classifier.
    \end{itemize}
    
    \subsection{Comparison}
	 Tables~\ref{tab:tab2}, \ref{tab:tab3}, and~\ref{tab:tab4} compare
	 the AUC, Recall, and G-mean between the state-of-the-art methods
	 and the proposed method (HADR). In general, HADR outperforms
	 the state-of-the-art methods on all datasets. HADR is outperformed by DDAE 
	 on only two parts, one of which is the Pc1 recall. Since
	 the G-mean performance of the proposed method (HADR) is better than DDAE,
	 this phenomenon implies that HADR tends to strike a better balance between
	 majority and minority class, rather than identifying as many minority class
	 samples as possible. The other case is the abalone20 G-mean.
	 As the AUC in the proposed method is better, this is explained by the fact that
	 the majority voting mechanism takes $0.5$ as the positive/negative threshold. 
	 Thus it may be that the proposed method (HADR) would yield a better G-mean
	 if a more appropriate threshold were chosen. Note that HADR
	 outperforms DDAE significantly in terms of recall for Wine3vs5 and Mw1.
	 As shown in Table~\ref{tab:tab1}, the Wine3vs5 imbalance ratio 
	 implies there are $69$ data blocks and about $20$ samples in each data
	 block, but the number of features is $11$. In this dataset, the $69$
	 classifiers would easily overfit. A similar situation 
	 occurs in Mw1. Evidently, DDAE performs more poorly in this kind of
	 situation. However, HADR achieves strong recall in both
	 datasets, likely because both DR and EL of the MLPs alleviate the tendency to
	 overfit.
    
    \begin{table}[htbp]
    \caption{AUC results}
    \begin{center}
    \setlength{\tabcolsep}{2.5pt}
    \begin{tabular}{|c|c|c|c|c|c|c|}
    \hline
    \diagbox{\textbf{Dataset}}{\textbf{Method}} & IML & CAdaMEC & MWMOTE & RP & DDAE & HADR\\
    \hline
    Pc1          & 0.679 & 0.731 & 0.782 & 0.807 & 0.870 & \bf{0.900}\\
    Pc3          & 0.582 & 0.731 & 0.671 & 0.726 & 0.744 & \bf{0.770}\\
    Pc4          & 0.730 & 0.828 & 0.778 & 0.873 & 0.813 & \bf{0.921}\\
    Mw1          & 0.653 & 0.728 & 0.728 & 0.702 & 0.817 & \bf{0.847}\\
    Wine3vs5     & 0.500 & 0.500 & 0.490 & 0.557 & 0.620 & \bf{0.820}\\
    abalone9vs18 & 0.719 & 0.736 & 0.740 & 0.702 & 0.824 & \bf{0.950}\\
    abalone19    & 0.628 & 0.500 & 0.579 & 0.773 & 0.852 & \bf{0.900}\\
    abalone20    & 0.802 & 0.693 & 0.598 & 0.904 & 0.965 & \bf{0.990}\\
    \hline
    \end{tabular}
    \label{tab:tab2}
    \end{center}
    \end{table}
    
    \begin{table}[htbp]
    \caption{Recall results}
    \begin{center}
    \setlength{\tabcolsep}{2.5pt}
    \begin{tabular}{|c|c|c|c|c|c|c|}
    \hline
    \diagbox{\textbf{Dataset}}{\textbf{Method}} & IML & CAdaMEC & MWMOTE & RP & DDAE & HADR\\
    \hline
    Pc1          & 0.852 & 0.519 & 0.630 & 0.889 & \bf{0.963} & 0.913\\
    Pc3          & 0.510 & 0.612 & 0.490 & 0.735 & 0.735 & \bf{0.781}\\
    Pc4          & 0.814 & 0.780 & 0.678 & 0.881 & 0.932 & \bf{0.970}\\
    Mw1          & 0.500 & 0.625 & 0.625 & 0.750 & 0.750 & \bf{1.000}\\
    Wine3vs5     & 0.000 & 0.000 & 0.000 & 0.333 & 0.333 & \bf{1.000}\\
    abalone9vs18 & 0.600 & 0.500 & 0.500 & 0.600 & 0.700 & \bf{0.875}\\
    abalone19    & 0.667 & 0.000 & 0.167 & 0.833 & \bf{1.000} & \bf{1.000}\\
    abalone20    & 0.800 & 0.400 & 0.200 & \bf{1.000} & \bf{1.000} & \bf{1.000}\\
    \hline
    \end{tabular}
    \label{tab:tab3}
    \end{center}
    \end{table}
    
    \begin{table}[htbp]
    \caption{G-mean results}
    \begin{center}
    \setlength{\tabcolsep}{2.5pt}
    \begin{tabular}{|c|c|c|c|c|c|c|}
    \hline
    \diagbox{\textbf{Dataset}}{\textbf{Method}} & IML & CAdaMEC & MWMOTE & RP & DDAE & HADR\\
    \hline
    Pc1          & 0.657 & 0.700 & 0.767 & 0.803 & 0.819 & \bf{0.848}\\
    Pc3          & 0.578 & 0.721 & 0.646 & 0.726 & 0.743 & \bf{0.781}\\
    Pc4          & 0.725 & 0.826 & 0.772 & 0.873 & 0.804 & \bf{0.890}\\
    Mw1          & 0.635 & 0.721 & 0.721 & 0.701 & 0.815 & \bf{0.816}\\
    Wine3vs5     & 0.000 & 0.000 & 0.000 & 0.510 & 0.550 & \bf{0.778}\\
    abalone9vs18 & 0.709 & 0.697 & 0.700 & 0.695 & 0.814 & \bf{0.854}\\
    abalone19    & 0.626 & 0.000 & 0.407 & 0.771 & 0.839 & \bf{0.840}\\
    abalone20    & 0.802 & 0.628 & 0.446 & 0.904 & \bf{0.964} & 0.933\\
    \hline
    \end{tabular}
    \label{tab:tab4}
    \end{center}
    \end{table}
    
    \subsection{Ablation Study}
	 To evaluate the effectiveness of the three HADR components---data block 
	 construction (DBC), dimensionality reduction (DR), and ensemble learning (EL)---we
	 performed an ablation study of the model on two public datasets: Pc1 and
	 Wine3vs5. 

    \subsubsection{Without DBC}
	 We evaluated the model performance without DBC, as DBC is responsible for
	 generating balanced data blocks. Without DBC, the model is faced with only one
	 imbalanced classification problem, rather than several balanced problems.
	 Therefore, there is no ensemble learning, as there is only one set of
	 imbalanced data. Thus it performs MLP only once. In Figs.~\ref{fig:abla1} 
	 and~\ref{fig:abla2} we refer to this situation as ``DR+MLP''. For Pc1
	 and Wine3vs5, the DR+MLP recall is zero, which means that the
	 model predicts no samples from a minority at all. The model easily
	 mispredicts samples as the majority class due to the heavy imbalance.

    \subsubsection{Without DR}
	 We also evaluated the model performance without DR, which means the dataset 
	 is split into multiple balanced blocks. Each block performs MLP training
	 and the prediction is generated by a majority vote of the multiple MLP
	 results.   
	 For models without dimensionality reduction, we can
	 determine that model with DBC yields a certain
	 level of prediction for minority class samples. However, without the DR
	 component, data distribution and noise are caused by extraneous dimensions of the
	 dataset, which yields results that are inferior to those of HADR.

    \Figure[t!][width=0.4\textwidth]{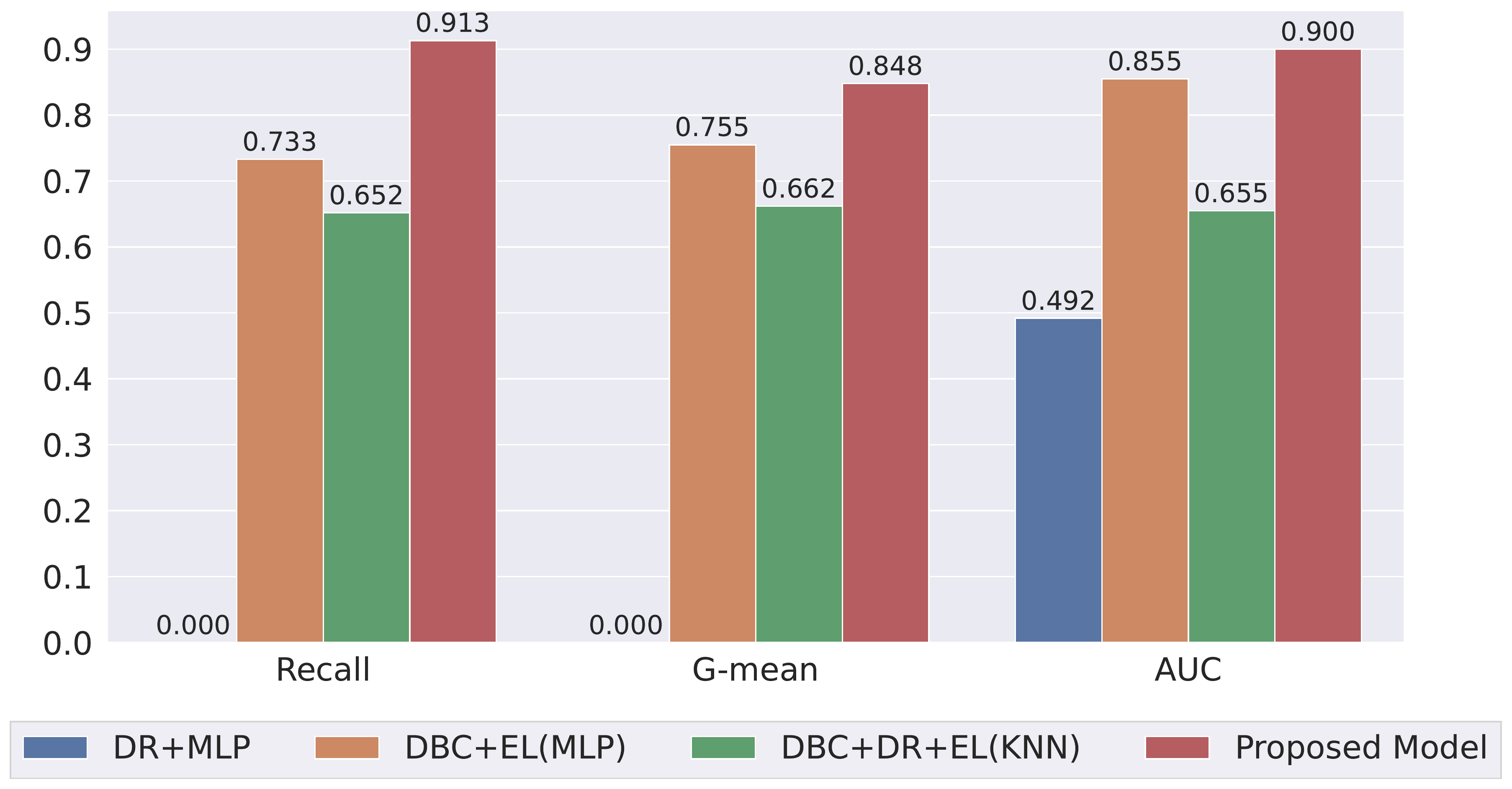}
    {Ablation study on PC1 dataset \label{fig:abla1}}
    \Figure[t!][width=0.4\textwidth]{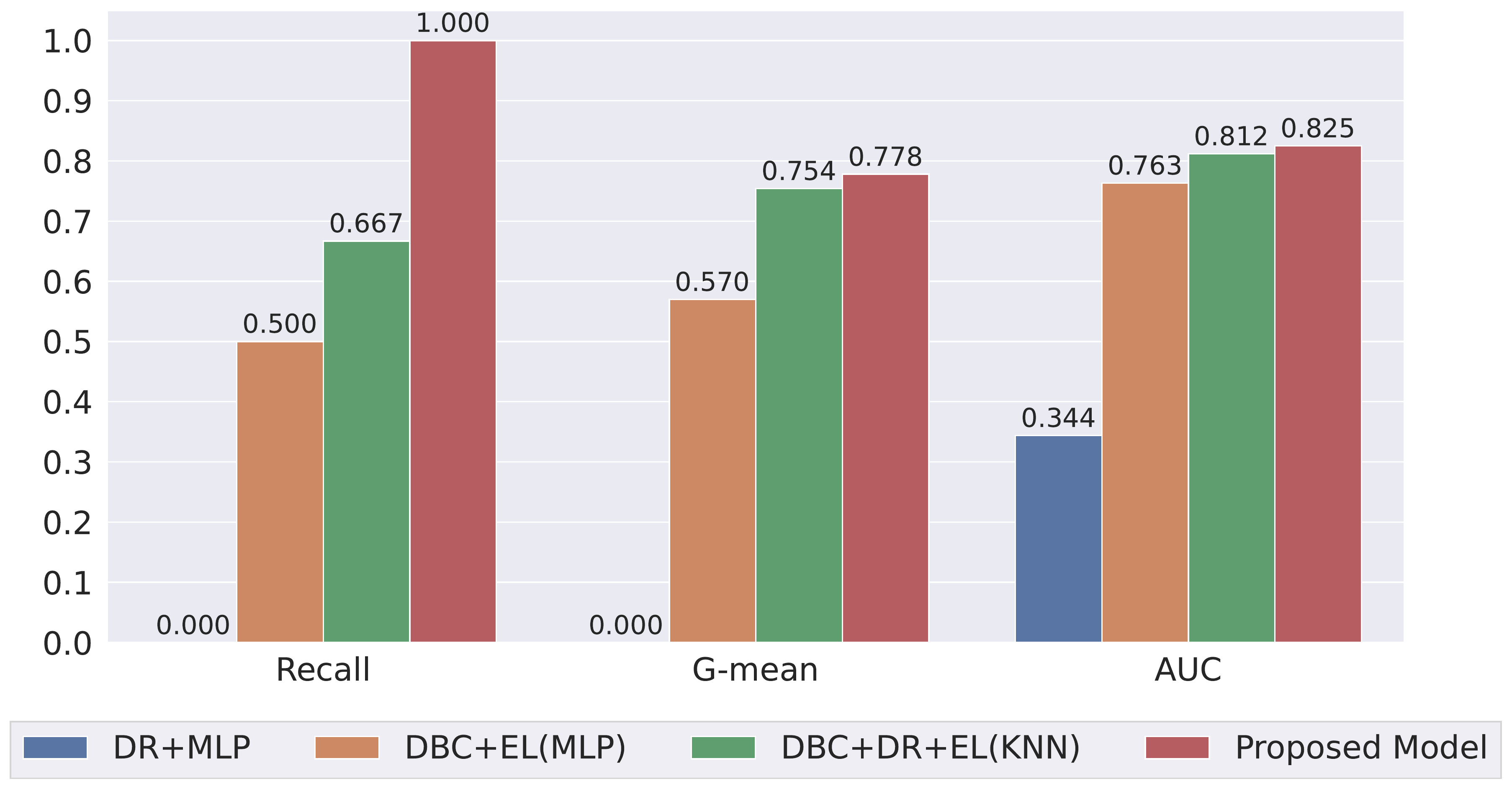}
    {Ablation study on Wine3vs5 dataset \label{fig:abla2}}

\section{Conclusion}
\label{sec:conclusion}
We propose a hybrid approach with dimension reduction (HADR) for binary
imbalanced dataset classification. This model contains components for data block
construction (DBC), dimensionality reduction (DR), and
ensemble learning (EL). The model outperforms state-of-the-art
models on several imbalanced datasets, which demonstrates that HADR is competitive
against existing model structures for imbalanced classification.

Currently, HADR is limited to binary imbalanced classification 
and is restricted to non-image data. We will consider multi-class
classification on imbalanced data, image data, or medical problems as future
work.

\EOD

\end{document}